\documentclass[10pt,fleqn]{scrartcl}
\usepackage[margin=20mm]{geometry}

\usepackage{authblk}
\usepackage{graphicx}

\title{CHA\textsubscript{2}: \underline{CH}emistry \underline{A}ware \underline{C}onvex \underline{H}ull \underline{A}utoencoder \\ Towards Inverse Molecular Design}

\author[1]{Mohammad Sajjad Ghaemi\thanks{MohammadSajjad.Ghaemi@nrc-cnrc.gc.ca}}
\author[1]{Hang Hu} 
\author[2]{Anguang Hu} 
\author[1]{Hsu Kiang Ooi}
\affil[1]{National Research Council Canada, Toronto, ON, Canada} 
\affil[2]{Suffield Research Centre, DRDC}
  
\begin{document}
\date{}
\maketitle
\begin{abstract} 
Optimizing molecular design and discovering novel chemical structures to meet certain objectives, such as quantitative estimates of the drug-likeness score (QEDs), is NP-hard due to the vast combinatorial design space of discrete molecular structures, which makes it near impossible to explore the entire search space comprehensively to exploit \emph{de novo} structures with properties of interest. To address this challenge, reducing the intractable search space into a lower-dimensional latent volume helps examine molecular candidates more feasibly via inverse design. Autoencoders are suitable deep learning techniques, equipped with an encoder that reduces the discrete molecular structure into a latent space and a decoder that inverts the search space back to the molecular design. The continuous property of the latent space, which characterizes the discrete chemical structures, provides a flexible representation for inverse design in order to discover novel molecules. However, exploring this latent space requires certain insights to generate new structures. We propose using a convex hall surrounding the top molecules in terms of high QEDs to ensnare a tight subspace in the latent representation as an efficient way to reveal novel molecules with high QEDs. We demonstrate the effectiveness of our suggested method by using the QM9 as a training dataset along with the Self-Referencing Embedded Strings (SELFIES) representation to calibrate the autoencoder in order to carry out the Inverse molecular design that leads to unfold novel chemical structure. \\
\end{abstract}

\section{Introduction}
\label{intro}
The rise of artificial intelligence (AI), in particular, emergence of generative machine learning (ML) techniques as a disruptive breakthrough technology, has revolutionized numerous fields, including molecular, material, and drug design, discovery, and optimization \cite{menon2022generative, 9756593}. As such taming an intractable combinatorial solution region in order to search for novel structure with desirable chemical properties has become a promising perspective in recent years \cite{blaschke2018application}. Additionally, data-driven inverse design inspired by machine learning methodologies is an effective approach to disseminate molecular knowledge from the training data to identify new candidates for chemical structures that meet specific properties \cite{sanchez2018inverse}. 

While typical machine learning methods are designed to exploit domain knowledge through training data to uncover the underlying distribution of data, novel structures are usually hidden in unexplored, low-density regions. As such, questing unprecedented chemical molecules requires deploying unique learning schemes to seek uninvestigated regions, which are restricted by the learned clues from existing information, such as chemical properties of matter \cite{menon2022generative}. To this end, learning a continuous bijective function to project the discrete structure of molecules into a lower-dimensional space and simultaneously map a random data point from the same lower-dimensional space onto a molecule representation is necessary \cite{Romez-Bombarelli2018}.

The ultimate goal of a high-dimensional data representation such as autoencoder is to find an underlying low-dimensional manifold that preserves the vital information specific to the desired properties, e.g. QEDs in the molecular design context. Thus, for separable distinct clusters of information, a well-decomposed group of salient signals in the latent space is expected from a perfect autoencoder model to account for sources of variation in the data \cite{joswig2020geometric}. As such, a convex combination of signals in the vicinity of data points with distinguished properties presumably entails a mixture of features reflecting prominent holistic patterns.

In this study, we propose a convex hull approach to generate out-of-distribution (OOD) novel molecules through the uniform samples restricted to the boundaries of high QEDs molecules in the latent space of autoencoder trained on QM9 data \cite{Ramakrishnan2014} via 1-hot-encoding of SELFIES representation.

\section{Relevant Prior Works}
\label{relevant}
In recent years, various molecular data encoding techniques, such as geometrical distance, molecular graph, and string specifications, have been employed to feed molecules as inputs to ML models for generating synthetic molecules. Among these techniques, string specifications, including simplified molecular-input line-entry system (SMILES), deep SMILES, and SELFIES, have gained significant attention due to their promising results \cite{menon2022generative, Romez-Bombarelli2018}. To effectively carry out sequential training, natural language processing (NLP) techniques, particularly Recurrent Neural Networks (RNN) and its variants such as Long Short-Term Memory (LSTM) and Gated Recurrent Unit (GRU), have been used successfully \cite{Ghaemi2022Generative}. Similarly, data compression and dimension reduction techniques, such as autoencoders and their generative counterpart, variational autoencoders (VAE) \cite{Kingma2014}, have been adopted for the same purpose by leveraging the holistic string representation as opposed to the sequential approach \cite{Romez-Bombarelli2018}.

In this study, we developed a novel type of autoencoder model that incorporates domain knowledge, specifically the QEDs, as the hallmarks of the Quantum Mechanics Database (QM9) data. The proposed CHA\textsubscript{2} method guides the autoencoder through high QEDs molecules in the latent space by constructing a convex hull entailing the points corresponding to the high QEDs molecules. Synthetic molecules are generated by taking random data points that are uniformly sampled from the boundaries of the convex hull surrounding the selected hallmarks of the latent space to the region of interest.

\section{Proposed Method and Experimental Design}
\label{method}
\begin{figure}
    \centering
    \includegraphics[scale=0.73]{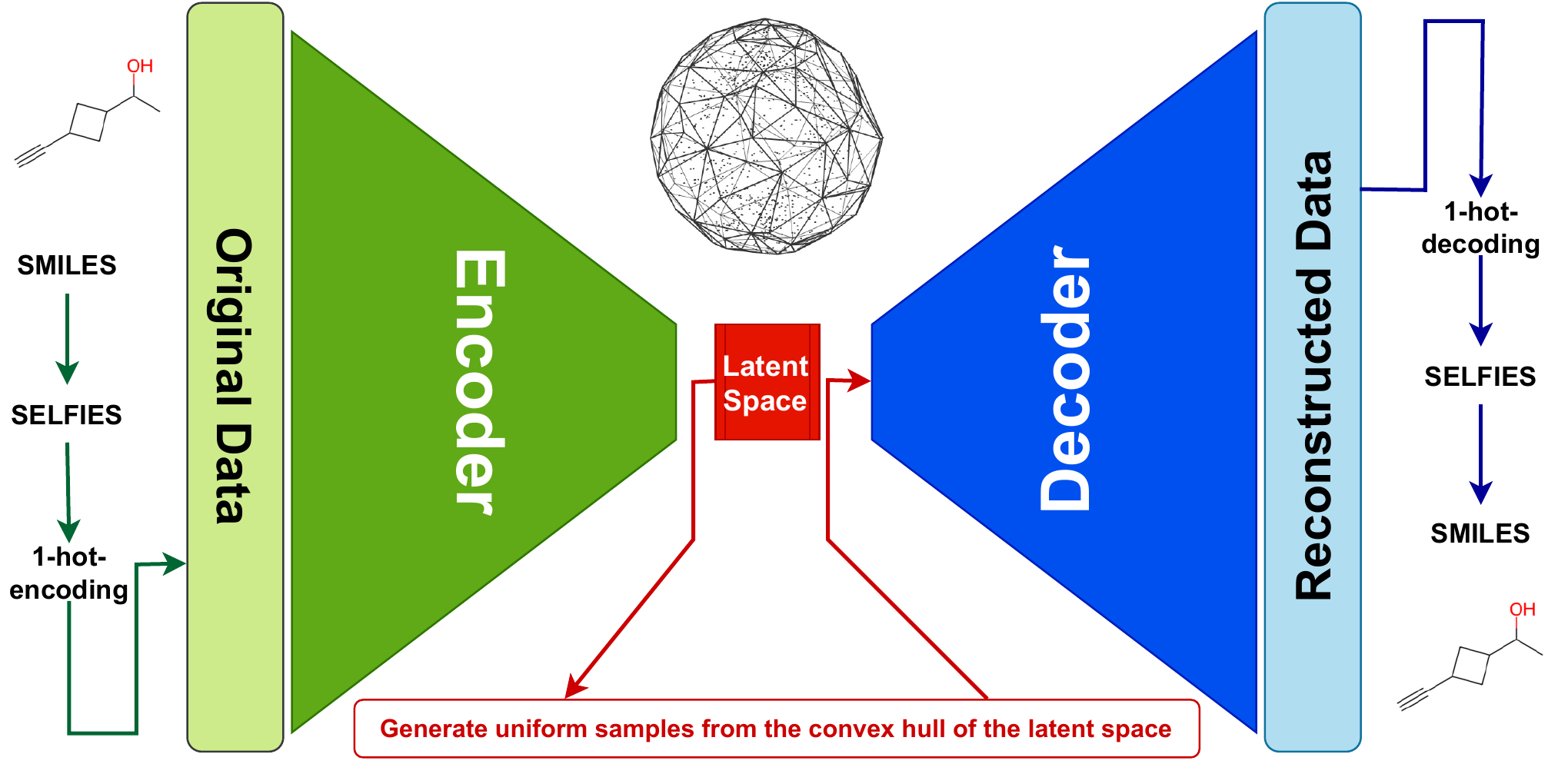} 
    \caption{Schematic of the CHA\textsubscript{2} approach for inverse molecular design, based on the convex hull associated with the top QEDs molecules. Discrete molecular structures represented by 1-hot-encoding of SELFIES and fed to the encoder network to form the continuous latent space. Consequently, decoder module generates synthetic molecules by sampling random points from the convex hull.}
    \label{fig:CHAE}
\end{figure}
In many cases, generative models are trained to estimate the probability distribution of the source data, with the expectation that the learned distribution will produce samples that share the same statistical properties as the real data. This approach facilitates the creation of a model capable of efficiently replicating data points that mimic the source data. However, for specific tasks such as drug discovery and material design, the generation of OOD synthetic data that lie beyond the original data distribution is crucial.

\begin{figure}
    \centering
    \includegraphics[width=\textwidth]{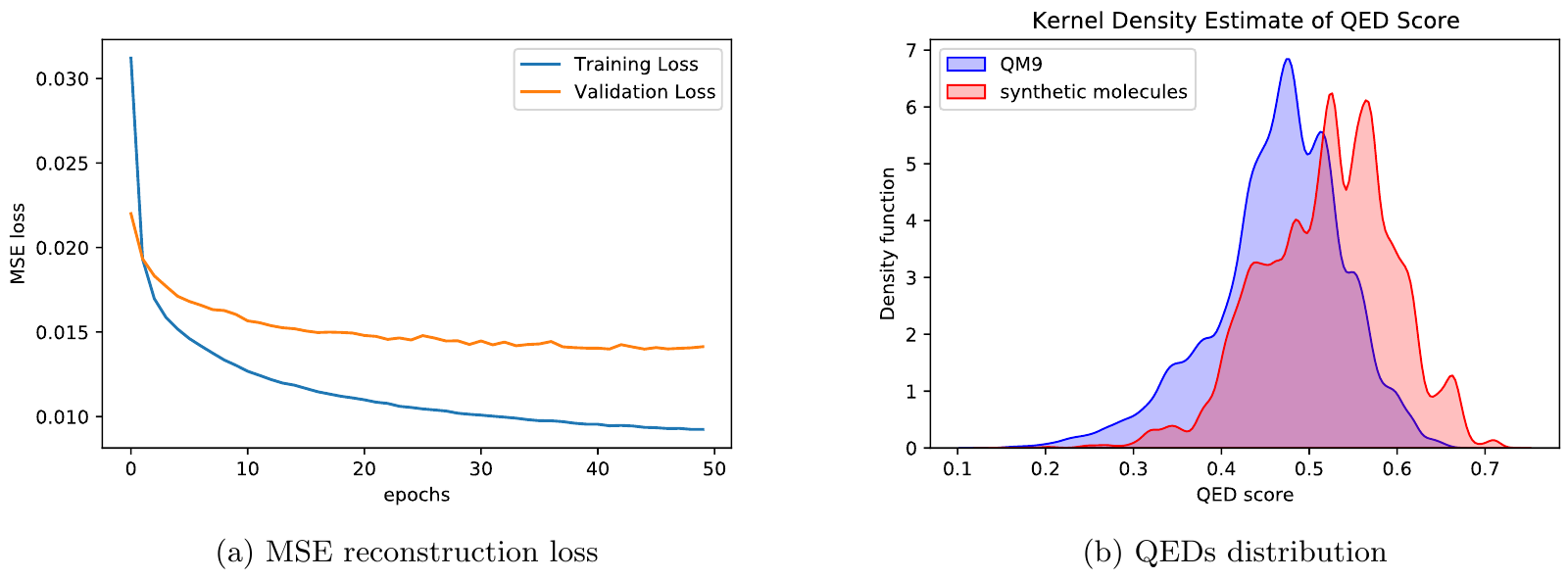}
 
    \caption{Results of CHA\textsubscript{2}. (a) Mean squared error (MSE) reconstruction loss of autoencoder training and validation. The blue line shows the training loss, while the orange line shows the validation loss. (b) The distribution of QEDs for the QM9 dataset and the synthetically generated molecules. The QEDs range from $0$ (least drug-like) to $1$ (most drug-like).}
    \label{fig:MSE-dist}
\end{figure}

\begin{figure}
    \includegraphics[width=\textwidth]{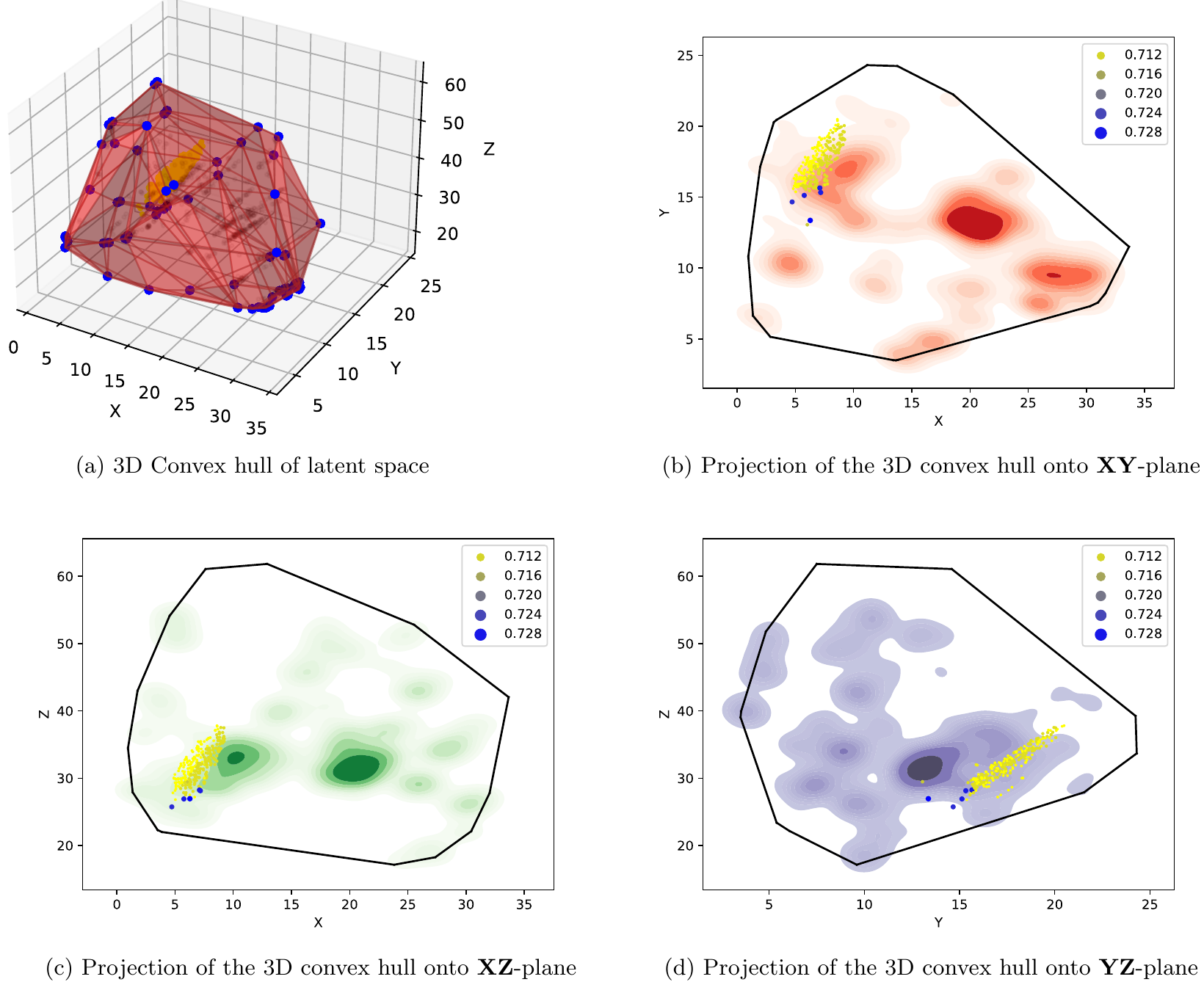}
    
    \caption{3D convex hull of the autoencoder latent space along with its projection onto the $\mathbf{XY}$, $\mathbf{XZ}$, and $\mathbf{YZ}$ coordinate planes. (a) Solid blue dots indicate the top QEDs molecules, the yellow region shows the synthetic molecules with QEDs $>0.7$, while the red area represents the space of entire random samples generated from the convex hull. (b-d) The yellow-to-blue spectrum of the QEDs is shown for synthetic molecules within the density of QM9 molecules. The QEDs range from yellow (low values) to blue (high values), indicating the distribution of QEDs within the polyhedron.}
    \label{fig:CHPr}
\end{figure}

The proposed CHA\textsubscript{2} method is novel in its approach to guiding the autoencoder towards high QEDs molecules in the latent space. This is accomplished by constructing a convex hull containing the points corresponding to the high QEDs molecules. Uniformly random samples are then generated, restricted to the boundaries of the convex hull, to empower the decoder to achieve high QEDs synthetic molecules. The autoencoder was trained using the 1-hot-encoding of SELFIES representation based on the QM9 data (as shown in Figure \ref{fig:CHAE}). Training data consisted of molecules with QEDs greater than $0.5$, while molecules with QEDs between $0.4$ and $0.5$ were used as validation data. During the molecular training and generating process, the molecule size was fixed at $19$ elements of SELFIES, which matched the maximum length of molecules in the training and validation data. The deep learning model was optimized using mean squared error (MSE) for autoencoder implementation in TensorFlow, and consisted of an array of layers comprising $ (250, 120, 60, 30, 8, 3, 8, 30, 60, 120, 250 ) $ neurons with rectified linear unit (ReLU) activation functions (except for the last layer, which used a sigmoid activation function) \cite{abadi2016tensorflow}. Figure \ref{fig:MSE-dist}a depicts the convergence of the MSE of the reconstruction loss on the training and validation data.
After training the autoencoder, a set of $100,000$ random samples was generated from the convex hull fit over top molecules with QEDs greater than $0.65$ in the latent space. Following the removal of duplicates and a sanity check, approximately $11,000$ valid synthetic molecules were obtained using this approach. The distribution of QEDs for both the QM9 dataset and the synthetically generated molecules is presented in Figure \ref{fig:MSE-dist}b. The 3D convex hull of the autoencoder latent space, as well as its projection onto the $\mathbf{XY}$, $\mathbf{XZ}$, and $\mathbf{YZ}$ coordinate planes, are illustrated in Figure \ref{fig:CHPr}. The solid blue dots indicate the top molecules with high QEDs, while the yellow region represents the space of synthetic molecules with QEDs greater than $0.7$, and the red area denotes the space of all random samples generated from the convex hull shown in Figure \ref{fig:CHPr}a. The yellow-to-blue spectrum of QEDs is displayed for the synthetic molecules within the density of QM9 molecules. This spectrum ranges from yellow for low QEDs to blue for high QEDs, illustrating the distribution of QEDs within the polyhedron, as depicted in Figures 3b to 3d.

\section{Results}
\label{results}
\begin{figure}
    \centering
    \includegraphics[scale=0.55]{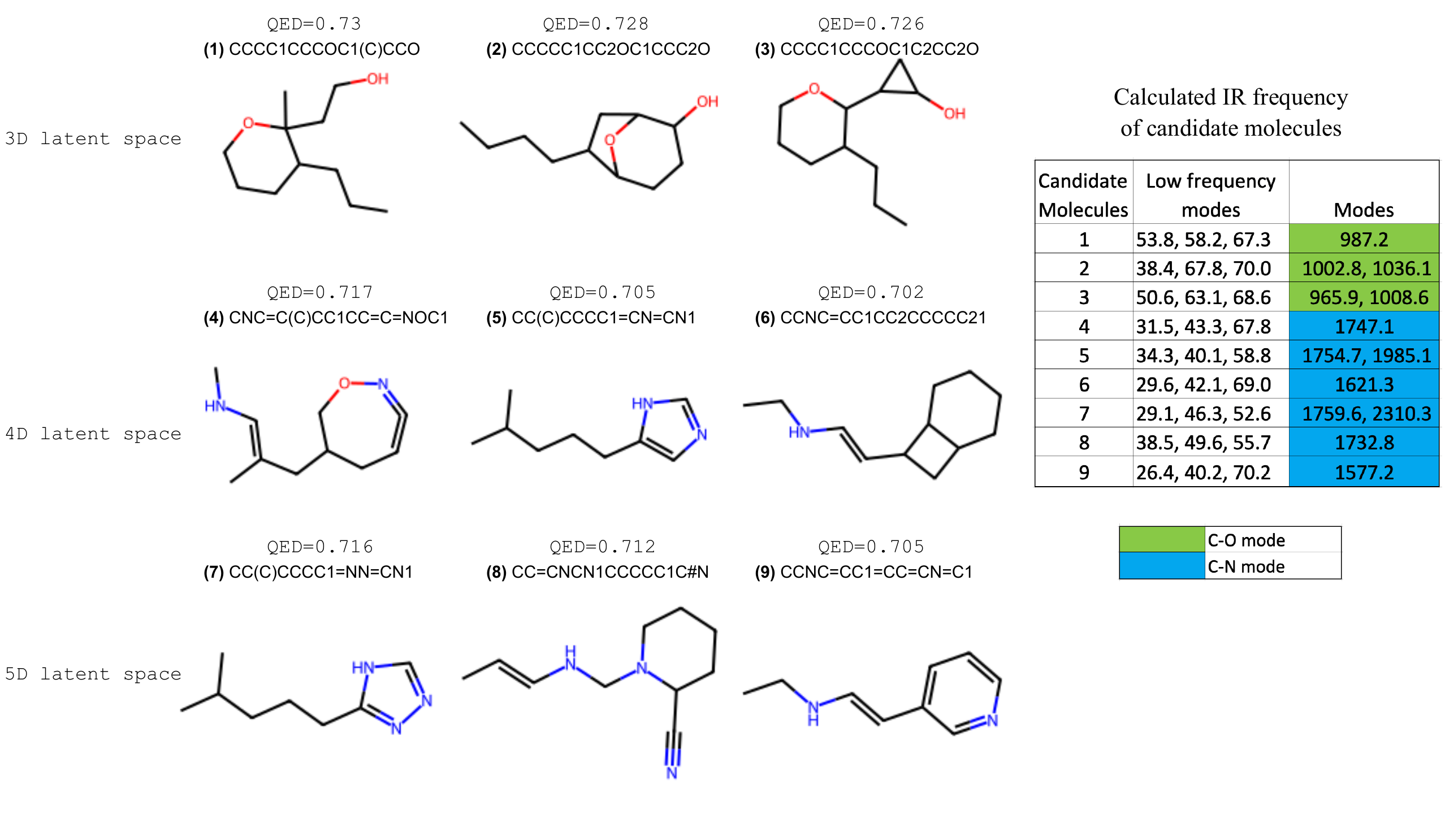} 
    \caption{Molecules with Higher QEDs in the Latent Space. Varying the number of units in the latent space resulted in additional molecular designs with higher QEDs. The top row shows the three (1-3) molecules with the highest QEDs generated from the CHA\textsubscript{2}. The table displays the calculated IR frequencies for these candidate molecules, revealing an interesting design principle. The unit for IR frequency is cm\textsuperscript{-1}.}
    \label{fig:syntheticmols}
\end{figure}

We explored the framework for design within three, four, and five dimensional latent spaces. Figure \ref{fig:syntheticmols} represents the best-in-class molecules based on QEDs for each respective latent space. We further calculated the generated molecules' vibration modes and infrared (IR) spectrum as a sanity check. Our findings indicate that all synthesized molecules have a positive frequency for their vibration modes, which confirms their physical stability as chemically viable compounds.

The structure relaxation and vibration mode are calculated using Density-functional theory with B3LYP exchange-correlation functional and 631G basis set, implemented under the Gaussian 16 simulation package\cite{g16, b3lyp, Becke, pople_basis}. The frequency of selected IR active modes is tabulated in Figure 4. In the case of molecules generated using the 3-dimensional latent space, the corresponding QEDs (candidates 1-3) were found to be superior to those generated using other dimensions (candidates 4-9). One design principle observed is that generated molecules with a C-O mode tend to result in higher QEDs.

Important physiochemical properties of the generated structures were also calculated and compared to the QEDs. These properties includes toxicity, intrinsic clearance, molecular weights and etc. We observed that the intrinsic clearance rate (CLint\_pred) is higher in cases where the QEDs is high. This is a critical property that indicates how well a drug can be metabolized by the hepatocytes.

Next we focused on the three-dimensions latent space to identify further design principles that would enhance QEDs. By generating a total of $11000$ molecules, we observed $13$ molecules that exhibited QEDs greater than $0.70$. Notably, our analysis revealed an absence of C-N bonds in molecules with higher QED values. Conversely, molecules with QED values less than 0.69 typically contained C-N bonds in their structures. The IR spectra of these molecules exhibited C-N functional groups with activity around 1700 cm\textsuperscript{-1}, while C-O groups were active around 1000 cm\textsuperscript{-1}, as depicted in Figure 4. This observation suggests that the IR spectra could be employed as a potential screening tool for experimentally identifying drug-like molecules.
 
\section{Conclusion and Future Work} 
The proposed CHA\textsubscript{2} is shown as a successful approach for inverse molecular design where convex hull played a central role in generating novel molecules from the autoencoder's latent space. The vibration modes and IR spectra analysis of the synthetic molecules showed that \emph{de novo} molecular structures are physically stable and chemically viable. While our analysis revealed a remarkable design principle for enhancing QEDs, namely the presence of chemical structures with C-O modes, we also observed that lack of C-N bonds led to higher QEDs. This finding suggests that the frequency of C-O bonds in the absence of C-N bonds in molecular structures may serve as an effective strategy for improving QEDs.

Additionally, with the rapid advancement of scalable quantum technology, a promising direction for CHA\textsubscript{2} to capitalize on the potential benefits of a hybrid approach that integrates machine learning and quantum chemistry simulation. In particular, the utilization of near-term Noisy Intermediate Scale Quantum (NISQ) technology has the potential to unlock new opportunities for tackling complex challenges in chemistry and materials design. This involve developing more sophisticated ML algorithms inspired by quantum chemistry and generative models inspired by transformers that leverage quantum computing hardware and deploy enriched datasets to enhance accuracy and performance.

\section*{Acknowledgements} This project is supported by the National Research Council Canada (NRC) under the AI for
Design Challenge Program and the Defence Research and Development Canada (DRDC).

\bibliographystyle{abbrv}
\bibliography{ref}

\end{document}